\documentclass[letterpaper, 10 pt, conference]{ieeeconf}  

\IEEEoverridecommandlockouts                              

\usepackage{graphics} 
\usepackage{epsfig} 
\usepackage{mathptmx} 
\usepackage{times} 
\usepackage{amsmath} 
\usepackage{amssymb}  
\usepackage{color}
\usepackage{booktabs}
\usepackage{subfigure}
\usepackage{multirow}
\usepackage{algorithm}
\usepackage{algorithmicx}
\usepackage{algpseudocode}
\usepackage{bm}
\usepackage{xcolor}
\usepackage{pifont}

\usepackage{balance}
\usepackage{colortbl}
\usepackage{marvosym}
\usepackage{ifsym}

\usepackage[
	colorlinks,
	anchorcolor=blue,
	linkcolor=cyan,
	urlcolor=red,
	citecolor=green,
	backref=page]{hyperref}

\title{\LARGE \bf 
IROAM: Improving Roadside Monocular 3D Object Detection \\ Learning from Autonomous Vehicle Data Domain
}

\author{Zhe Wang$^{1 \dagger}$, Xiaoliang Huo$^{2 \dagger}$, Siqi Fan$^{1}$, Jingjing Liu$^{1}$, Ya-Qin Zhang$^{1}$, Yan Wang$^{1}$\textsuperscript{\Letter}
\thanks{$^{1}$Zhe Wang, Siqi Fan, Yan Wang\textsuperscript{\Letter}, Jingjing Liu, Ya-Qin Zhang\textsuperscript{\Letter} are with the Institute for AI Industry Research (AIR), Tsinghua University, Beijing, China.
{\tt\small \{wangzhe, fansiqi, wangyan, zhangyaqin \}@air.tsinghua.edu.cn}}
\thanks{$^{2}$Xiaoliang Huo is with the School of Software, Beihang University, Beijing, China.}
\thanks{$^{\dagger}$Contributed equally to this work, \textsuperscript{\Letter} is the correspondence author.}
}

\begin{document}

\maketitle
\thispagestyle{empty}
\pagestyle{plain}

\begin{abstract}

In autonomous driving, The perception capabilities of the ego-vehicle can be improved with roadside sensors, which can provide a holistic view of the environment. However, existing monocular detection methods designed for vehicle cameras are not suitable for roadside cameras due to viewpoint domain gaps. To bridge this gap and \textit{Improve ROAdside Monocular 3D object detection}, we propose IROAM, a semantic-geometry decoupled contrastive learning framework, which takes vehicle-side and roadside data as input simultaneously. IROAM has two significant modules. In-Domain Query Interaction module utilizes a transformer to learn content and depth information for each domain and outputs object queries. Cross-Domain Query Enhancement To learn better feature representations from two domains, Cross-Domain Query Enhancement decouples queries into semantic and geometry parts and only the former is used for contrastive learning. Experiments demonstrate the effectiveness of IROAM in improving roadside detector's performance. The results validate that IROAM has the capabilities to learn cross-domain information.

\end{abstract}

\section{INTRODUCTION}

The primary challenge in autonomous driving lies in achieving accurate detection of the surroundings. However, recent research targets enhancing the perception ability of ego-vehicle, which is constrained by sensor limitations in certain scenarios. For instance, cameras mounted on vehicles may suffer from occlusions and restricted visibility. Conversely, roadside cameras, positioned at higher elevations, offer a holistic view of the environment. Thus, there's a growing focus on leveraging roadside sensors, like cameras, to overcome these perception challenges.

\begin{figure}[t]
	\centering  
	\includegraphics[width=0.9\linewidth]{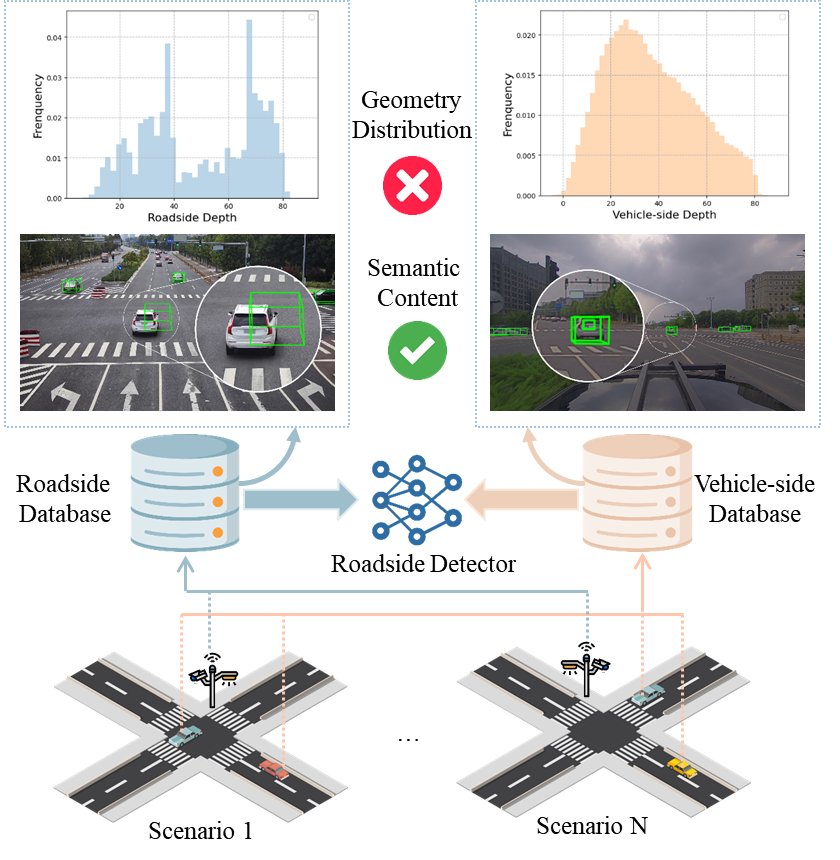} 
	\caption{Vehicle-side and roadside data have view domain gaps. The same vehicle captured in vehicle-side and roadside images has similar semantic content with slight differences in viewpoint. However, geometry distributions of objects (like depth distribution) varies significantly between two data domains.
}  
	\label{Fig:Distribution Difference}   
\end{figure}

 With the lack of precise depth information, 3D object detection using a monocular camera is an ill-posed problem. Most monocular detection methods are designed for vehicle-side images and have inferior performances when directly applied to roadside images. This is because there is a view domain gap between these two side images, which can be seen from distribution differences in the two data domains (shown in Figure~\ref{Fig:Distribution Difference}).

 There are several datasets~\cite{yu2022dairv2x,v2x-seq,ye2022rope3d} that provide benchmarks for roadside 3D object detection, such as BEVHeight~\cite{yang2023bevheight}. These approaches are specified for roadside and obtain better performance when trained on more roadside images. However, the process of calibration, collecting, and annotation of roadside datasets is time-consuming, which limits them from being extended to large-scale models. Since there are plenty of vehicle-side datasets like~\cite{Geiger2013KITTI,nuscenes2019,waymo}, it is promising to leverage vehicle-side data to assist the model in learning better feature representations from roadside data. One intuitive approach is to train models on a combination of vehicle-side and roadside data. However, the view domain gap between these two types of datasets (shown in Figure~\ref{Fig:Distribution Difference}) will hinder the model's performance and increase the difficulty of model convergence.

To address the challenges of learning from diverse view domains, contrastive learning provides a powerful tool, enabling models to extract meaningful representations by enhancing the similarity among akin instances while diminishing it among unlike ones across two data domains. This necessitates a pioneering contrastive learning approach tailored for bridging the view domain gap, focusing on identifying the parallels and variances between datasets from different view domains. As illustrated in Figure~\ref{Fig:Distribution Difference}, an identical object captured in both vehicle-side and roadside images shares similar semantic content despite slight differences in viewpoint. However, the depth distribution of objects relative to the camera positions, as showcased in Figure~\ref{Fig:Distribution Difference}, varies significantly between the roadside and vehicle-side datasets. Therefore, these two data domains share semantic similarities, with the primary distinction lying in the geometric distribution of objects.


In this paper, we propose a semantic-geometry decoupled contrastive learning framework IROAM that leverages tremendous and diverse vehicle-side data to improve the performance of roadside object detection models.  It utilizes a DETR-based detector to extract object-centric query features, which are designed for cross-view contrastive learning between vehicle-side and roadside data. The whole framework of IROAM can be divided into Feature Encoder, In-Domain Query Interaction, and Cross-domain Query Enhancement. First, Feature Encoder extracts content features and depth features from the raw image. Then, a set of object queries is initialized and updated by interacting with content features and depth features in the In-Domain Query Interaction. In this process, object queries learn the information within a single data domain (roadside or vehicle-side) and can be used to predict corresponding objects
through several specific heads. Finally, each object query will be decoupled into semantic and geometry parts in Cross-Domain Query Enhancement and only the former is used for contrastive learning so that the resource data domain (vehicle-side) can help the model to learn a better representation for detection in the target data domain (roadside). The semantic part contains the semantic similarity of two view data domains, which can be used to calculate contrastive loss and classification loss. The geometry part represents their geometry differences, which are supervised by the ground truth of two datasets respectively to calculate regression loss. Our contributions can be summarized as follows:


\begin{itemize}
    \item We propose IROAM, a novel semantic-geometry decoupled contrastive learning framework for vehicle and roadside data domains, which aims to enhance roadside monocular detection performance with the help of the vehicle data domain.
    \item We extract object-centric query features with the DETR-based detector and decouple them into semantic and geometry parts to learn a better representation for detection.
    \item We conduct experiments using different vehicle-side and roadside datasets, demonstrating that IROAM enhances detection performance for roadside detectors and possesses the ability to cross-domain learning.
\end{itemize}

\section{Related Work}
\subsection{Monocular 3D detection}

Traditional monocular detectors follow a center-guided approach where they process a single input image and focus on the centre point of the target. This pipeline aligns with the established methods used in traditional 2D detectors~\cite{ren2017fasterrcnn, lin2020focalloss, tian2019fcos, zhou2019objects}. However, a recent and influential trend within the field involves the adoption of transformer~\cite{vaswani2017attention} for the reconfiguration of detection tasks~\cite{carion2020detr, zhu2021deformabledetr}. DETR~\cite{carion2020detr} employs a set of object queries that yield detection results directly through cross-attention decoding. Similarly, Deformable DETR~\cite{zhu2021deformabledetr} introduces deformable attention mechanisms, resulting in a substantial reduction in training time.

Diverging from previous approaches, MonoDETR~\cite{zhang2022monodetr} presents an innovative departure by employing 3D object candidates as queries. Moreover, it introduces an attention-based depth encoder designed for encoding depth embeddings. This depth-guided pipeline allows the encoder to harness attention mechanisms to dynamically explore depth cues originating from depth-related regions within a global context. Impressively, MonoDETR attains superior performance while minimizing 3D-specific inductive biases. In this work, we choose MonoDETR as our monocular 3D detection backbone and enhance its performance using contrastive learning for cross-view data domains.

\subsection{Contrastive learning}

Contrastive learning has emerged as a powerful method in the field of computer vision, enabling self-supervised feature learning without heavily relying on extensive labeled datasets~\cite{chen2020simclr, Wei2021soco, li2022simipu, chen20224dcontrast}. This methodology involves aligning similar data points closely in an embedding space while distancing dissimilar ones, facilitating the recognition of significant patterns.  Contrastive pre-training has shown significant achievements across various image perspectives within the image domain~\cite{he2020momentum, tian2020contrastivemul, caron2020unsuplearning, grill2020bootstrap, wang2021dense}. Recent advancements, notably CO\^{}3~\cite{chen2022co3}, has employed contrastive learning to improve 3D detection using point clouds from both vehicle-side and roadside views, demonstrating its effectiveness and robustness in bridging the domain gaps between two sides. Therefore, further investigation into contrastive learning for cross-view data domains is crucial for tackling domain discrepancy between vehicle-side and roadside images.


\section{Method}


\begin{figure*}[ht]
	\centering  
	\includegraphics[width=0.85\linewidth]{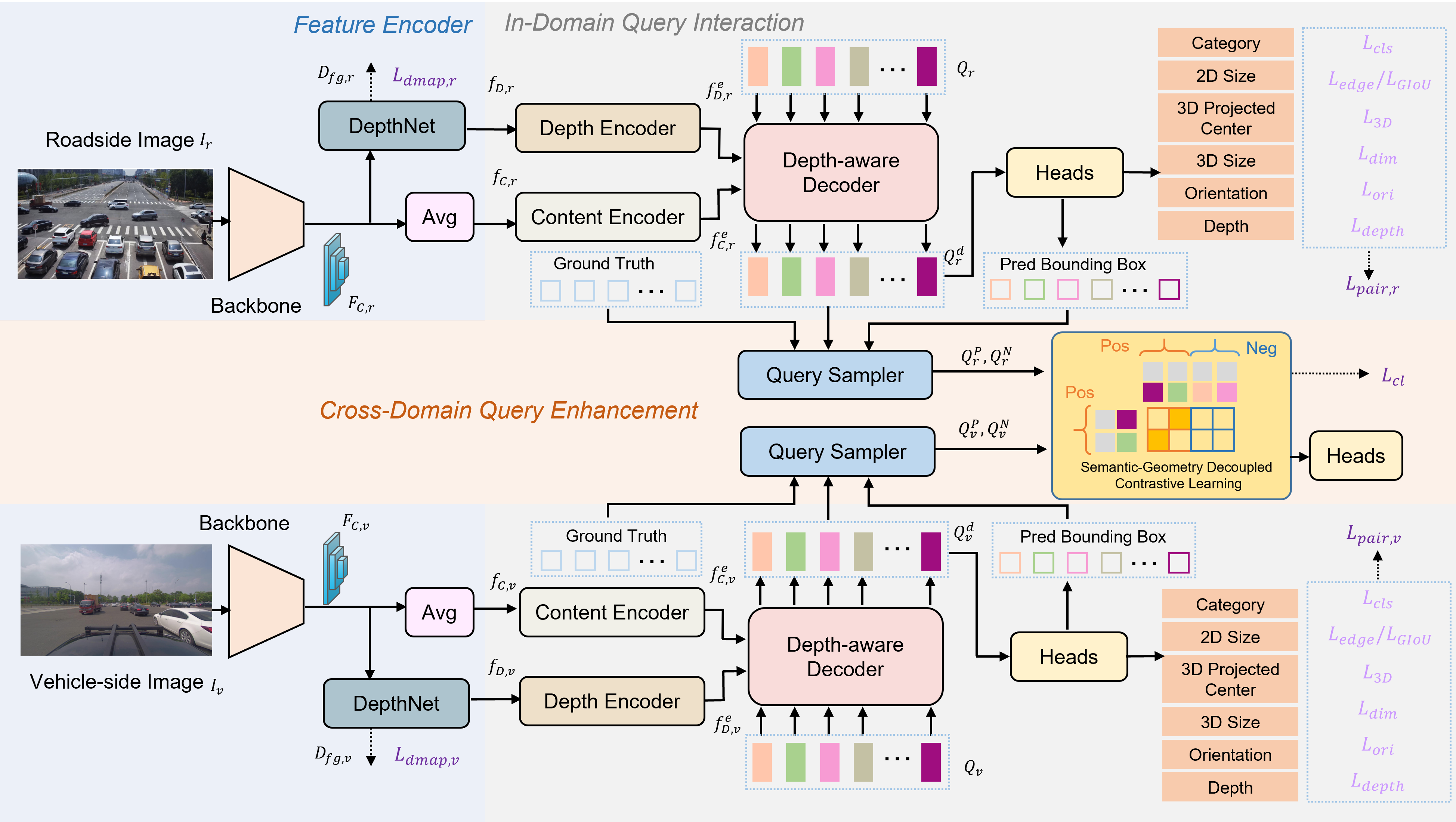} 
	\caption{The framework of IROAM contains a roadside branch and a vehicle-side branch and each branch has the same architecture of Feature Encoder and In-Domain Query Interaction module. $\ast \in \{r,v\}$determines variables belonging to roadside or vehicle-side. For each branch, multi-scale image features $F_{C,\ast}$ can be obtained from input image $I_{\ast}$. They can be transformed into content feature $f_{C,\ast}$ and depth feature $f_{D,\ast}$. The DepthNet predicts a foreground depth map $D_{fg,\ast}$ and supervise it with ground-truth depth.  $Q_{ \ast }$ can adaptively aggregate features from content embedding $f^{e}_{C,\ast}$ and depth embeddings $f^{e}_{D,\ast}$ and be updated as $Q^{d}_{ \ast }$, Then it will be transformed into positive sample set $Q^{P}_{ \ast }$ and negative sample set $Q^{N}_{ \ast }$ through Query Sampler. Finally,  each object query will be decoupled into semantic and geometry parts and only the former is used for contrastive learning.}
	\label{fig:framework}   
\end{figure*}


In this section, we introduce our semantic-geometry decoupled contrastive learning framework for roadside monocular 3D object detection, named IROAM. It utilizes a DETR-based detector to extract query features, which can be enhanced for better performance through contrastive learning between vehicle-side and roadside data domains. The whole framework of IROAM can be divided into Feature Encoder, In-Domain Query Interaction, and Cross-domain Query Enhancement, as shown in Figure~\ref{fig:framework}. 


\subsection{Feature Encoder}

The Feature Encoder takes a raw image as input and outputs its content feature $f_{C}$ and depth feature $f_{D}$ via a depth predictor DepthNet. Following~\cite{zhang2022monodetr}, we extract multi-scale image features given input image $I\in \mathbb{R}^{H \times W \times 3}$ through image backbone like ResNet-50~\cite{he2016resnet}. The feature maps from the last three stages of the image backbone can be denoted as $F_{C,r}$ for roadside and $F_{C,v}$ for vehicle-side.

\begin{equation}
F_{C,\ast} = \left\{f^{\frac{1}{s}}_{C,\ast} \in \mathbb{R}^{\frac{H}{s} \times \frac{W}{s} \times C}, s \in (8,16,32), \ast \in \{r,v\} \right\}
\end{equation}

Take roadside as an example, features ($f^{\frac{1}{32}}_{C,r}$ and $f^{\frac{1}{8}}_{C,r}$) are first transformed into a unified scale like $f^{\frac{1}{16}}_{C,r}$ using nearest-neighbor interpolation and convolutional layer respectively. Then, they can be integrated by element-wise average to form content feature $f_{C,r}$. 

Depth features $f_{D.r} \in \mathbb{R}^{\frac{H}{16} \times \frac{W}{16} \times C} $ can be extracted from $F_{C,r}$ using two $3 \times 3$ convolutional layers. For depth supervision,  a foreground depth map $D_{fg,r} \in \mathbb{R}^{\frac{H}{16} \times \frac{W}{16} \times (D+1)}$ is generated from $f_{D,r}$ through a $3 \times 3$ convolutional layer.  We use Focal Loss to supervise $D_{fg,r}$ with element-wise ground-truth depth, denoted as  $L_{\text{dmap,r}}$. 

We can get content features $f_{C,v}$, depth features $f_{D.v}$, and depth map loss $L_{\text{dmap,v}}$ for vehicle-side following the same pipeline.





\subsection{In-Domain Query Interaction}
\label{chap:in_domain}

We use transformer architecture~\cite{vaswani2017attention} to conduct interactions between content/depth features and object queries within a single data domain. The whole process consists of a content encoder, a depth encoder, and a depth-aware decoder. 


To illustrate the details, we take roadside as an example, and the pipeline of vehicle-side is the same. First, the content and depth encoder independently transform the content feature and depth feature into embeddings with global receptive fields, denoted as $f_{C,r}^{e}$, $f_{D,r}^{e} \in \mathbb{R}^{\frac{HW}{16^2} \times C}$. We apply one encoder block for the depth encoder and three blocks for the content encoder. Each encoder block is composed of a self-attention layer and a feed-forward neural network (FFN).  We add sine/cosine positional embedding to$f_{C,r}^{e}$, $f_{D,r}^{e}$ to supplement the absent 2D spatial structure.

In the depth-aware decoder, Object queries $Q_{r} \in \mathbb{R}^{N \times C}$ can adaptively aggregate features from depth and content embeddings $f_{C,r}^{e}$, $f_{D,r}^{e}$. We utilize three depth-aware decoder blocks, comprising a series of components including depth cross-attention, self-attention, content cross-attention, and a FFN. This design aims to provide one object with effective depth information and semantic content. We also add a learnable depth positional encoding for depth embedding $f_{D,r}^{e}$ following ~\cite{zhang2022monodetr}. The output queries, which can be denoted as $Q_{r}^{d}$, are sent to specific heads and calculate losses for category prediction $L_{cls}$, 2D bounding box $L_{edge}$, $L_{GIoU}$, 3D center $L_{3D}$, 3D dimension $L_{dim}$, orientation $L_{ori}$ and depth $L_{depth}$ following~\cite{zhang2022monodetr}.







We utilize the Hungarian algorithm~\cite{carion2020detr} to match the orderless object queries with ground-truth labels. For each query-label pair, we formulate the matching cost as:
\begin{equation}
\begin{aligned}
C_{\text {match }}= & \lambda_1 \cdot L_{\text {cls }} +\lambda_2 \cdot L_{3D} \\
& +\lambda_3 \cdot L_{\text {edge }}+\lambda_4 \cdot L_{GIoU},
\end{aligned}
\end{equation}
For each matched pair in domain, we compute the loss based on the matching cost $C_{\text{match}}$ as:
\begin{equation}
\begin{aligned}
& L_{\text {pair }}=C_{\text {match }}+\lambda_5 \cdot L_{\text {dim }} \\
& \quad +\lambda_6 \cdot L_{\text {ori }}+\lambda_7 \cdot L_{\text {depth }}
\end{aligned}
\end{equation}
The loss of the matched pair in roadside and vehicle-side domains can be denoted as $L_{\text{pair,r}}$ and $L_{\text{pair,v}}$.

\subsection{Cross-Domain Query Enhancement}
\label{Section:CDQE}
To assist the model in learning better feature representations from data domains in different views, we introduce cross-domain query enhancement based on contrastive learning. 

\textbf{Query Sampler} 
Through In-Domain Query Interaction, we will get the query features $Q^{d}_{r}=\{q_{r}^{i} | i=1,...,N\}$, $Q^{d}_{v}=\{q_{v}^{i} | i=1,...,N\}$ ($N$ is the number of queries) obtained from the roadside and vehicle-side image respectively. We introduce Query Sampler to select positive and negative samples from $Q^{d}_{r}$, $Q^{d}_{v}$ for contrastive learning. 

Since the same real object can be represented with different queries during this process, we need to define which query should match this object. We first compute the loss introduced in Section ~\ref{chap:in_domain} for each query and predict corresponding bounding boxes, and then utilize Hungarian algorithm~\cite{carion2020detr} to match the unordered queries with ground-truth labels. During the Bipartite matching process, we obtain the linear assignment score matrix as the matching results. Therefore, for roadside domain, we can define $K_{r}$ queries that have the highest matching scores with ground-truth labels $GT_{r}$ ($K_{r}$ is the number of $GT_{r}$) as the positive samples set $Q^{P}_{r}=\{q^{P}_{r,i} | i=1,...,K_{r}\}$ and $K_{r}$ queries that have the lowest matching scores as the negative samples set $Q^{N}_{r} = \{q^{N}_{r,i} | i=1,...,K_{r}\}$. For vehicle-side domain, we can obtain positive and negative samples set $Q^{P}_{v} = \{q^{P}_{v,i} (i=1,...,K_{v})\}$ and $Q^{N}_{v} = \{q^{N}_{v,i} | i=1,...,K_{v}\}$ ($K_{v}$ is the number of ground-truth $GT_{v}$) as the same way.

To enhance the generality of the model, we merge the sample sets from roadside and vehicle-side together as $Q^{P} = \{Q^{P}_{r}, Q^{P}_{v}\}, Q^{N} = \{Q^{N}_{r}, Q^{N}_{v}\}\}$, and then compute the similarity between each of them for contrastive learning between different data domain.

\begin{figure}[htbp]
	\centering  
	\includegraphics[width=0.8\linewidth]{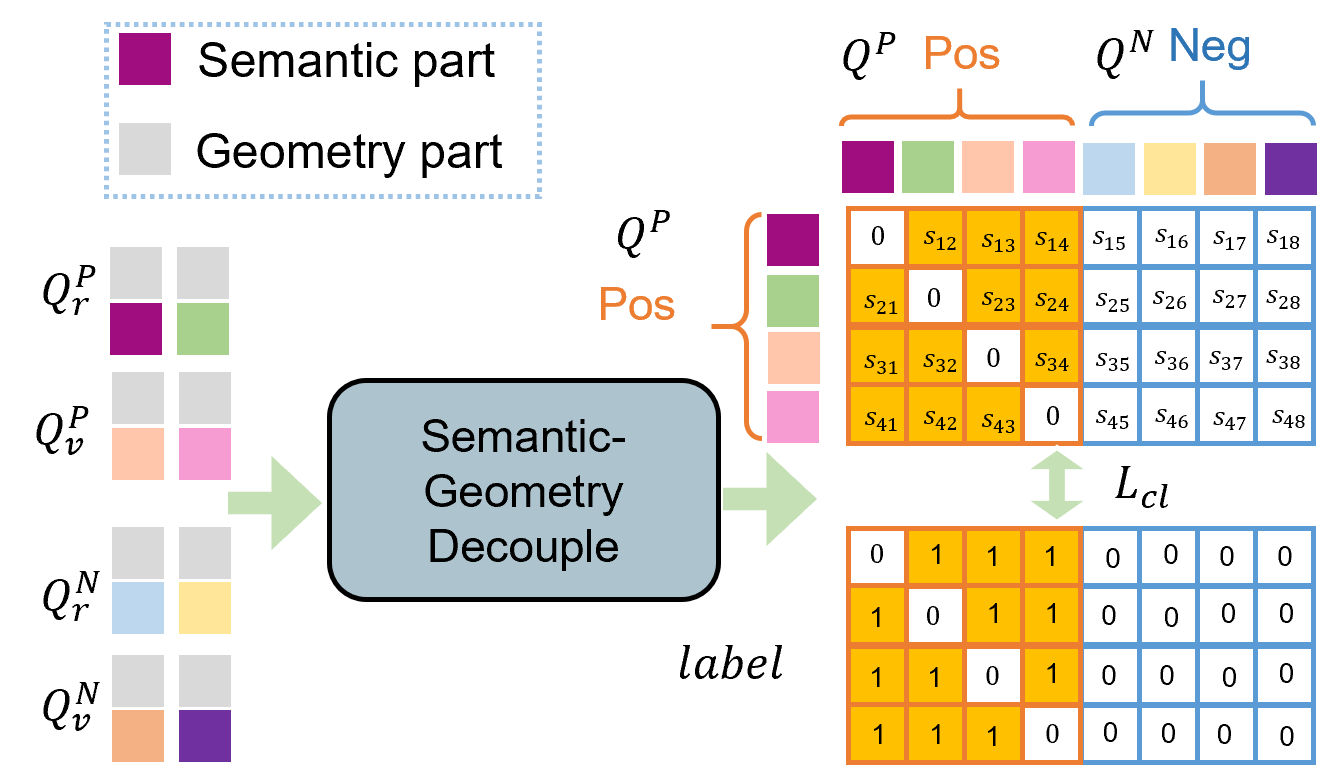} 
	\caption{The procedure of semantic-geometry decoupled Contrastive Learning. Every query is disentangled into a semantic feature (colored one) and a geometry feature (grey one) and the former are used to calculate $l_{cl}$}  
	\label{fig:CL}   
\end{figure}
 
\textbf{Semantic-Geometry Decoupled Contrastive Learning}
Since the goal of cross-domain enhancement mainly focuses on the semantic generality of objection detection instead of the geometric difference between different views, we split every single query feature $q\in1 \times N$ in positive or negative sample set into two parts (colored one and grey one in Figure~\ref{fig:CL}), the former $\frac{N}{2}$ channels can be denoted as semantic query $q^{sem} \in 1 \times \frac{N}{2}$, while the latter $\frac{N}{2}$ channels as geometric query $q^{geo} \in 1 \times \frac{N}{2}$. Then we apply the cross-domain contrastive learning introduced above only to $q^{sem}$. Both positive sample set $Q^{P}$ and negative sample set $Q^{N}$ have $K=K_{r}+K_{v}$ samples, and the similarity and similarity label between $i$-th query ($i=1,...,K$) in $Q^{P}$ and $j$-th query ($j=1,...,2K$) in set $(Q^{P},Q^{N})$ can defined as:

\begin{equation}
 s_{i,j} = 
\begin{cases}
  \text{sigmoid} \left( \frac{\langle q_{i}^{sem},q_{j}^{sem} \rangle}{\Vert q_{i}^{sem} \Vert_{2} \cdot \Vert q_{j}^{sem} \Vert_{2}}  \right), & i \neq j \\
 0, & i =j 
 \end{cases}
\end{equation}

\begin{equation}
label_{i,j}=
\begin{cases}
1& \text{$i,j \in P$ and $i \neq j$}\\
0& \text{otherwise}
\end{cases}
\end{equation}

$\langle \cdot \rangle$ means inner product. $P$ is the index set of $Q^{P}$. The overall contrastive loss $L_{cl}$ is defined as:
\begin{equation}
L_{cl} = \sum_{i=1}^{K} \sum_{j=1}^{2K}   \Vert s_{i,j} - lable_{i,j} \Vert_{1}
\end{equation}



\subsection{Overall Loss}

We can infer 3D bounding boxes directly from outputs, requiring no NMS post-processing. For training, we respectively compute the loss for each query-label pair and the contrastive loss between roadside and vehicle-side data domains. The overall loss with all $N$ queries for a pair of roadside and vehicle-side images as:

\begin{equation}
Loss=\frac{1}{K_{v}} \cdot \sum_{n=1}^{N} L_{\text {pair,v }}^n + \frac{1}{K_{r}} \cdot \sum_{n=1}^{N} L_{\text {pair,r }}^n +   L_{\text {dmap,r}}+  L_{\text {dmap,v}} +  L_{\text{cl}}
\end{equation}
where $K_{v}$ and $K_{r}$ denote the number of valid query-label pairs in the vehicle-side dataset and roadside dataset respectively. $L_{\text{dmap,r}}$ and $L_{\text{dmap,v}}$ represent the Focal loss for the predicted foreground depth map in the two data domain.

\section{Expermients}

\begin{figure*}[htbp]
	\centering  
	\includegraphics[width=0.85\linewidth]{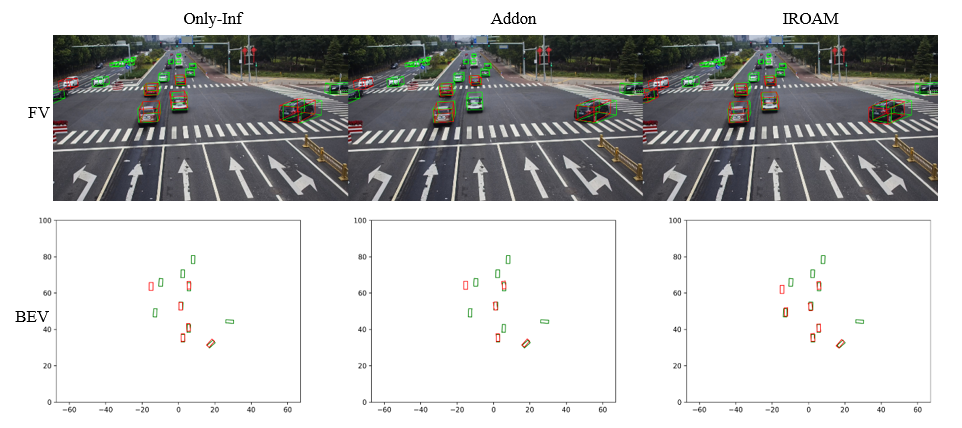} 
	\caption{Visualization results of IROAM. From BEV, it is clear that prediction bounding boxes (red) and labels (green) from IROAM are better aligned than Only-inf and Addon methods.}  
	\label{fig:vis_results}   
\end{figure*}
\subsection{Experimental Setup}
\begin{table*}[htbp]
  \footnotesize
  \centering
  \resizebox{\linewidth}{!}{
    \begin{tabular}{ccccccccccccccc}
    \hline
    \multirow{2}{*}{\textbf{Vehicle-side}} & \multirow{2}{*}{\underline{\textbf{Roadside(Eval)}}}& \multirow{2}{*}{\textbf{Method}} & \multicolumn{3}{c}{\textbf{$AP_{3D}(IoU=0.7)$}} & \multicolumn{3}{c}{\textbf{$AP_{BEV}(IoU=0.7)$}} & \multicolumn{3}{c}{\textbf{$AP_{3D}(IoU=0.5)$}} & \multicolumn{3}{c}{\textbf{$AP_{BEV}(IoU=0.5)$}}
     \\ \cline{4-15} 
     & & & Easy & Mod & Hard & Easy & Mod & Hard  & Easy & Mod & Hard & Easy & Mod & Hard \\ \hline
   \ding{55} & DAIR-V2X-I  & BEVHeight & 5.99 & 5.19 & 5.23 & 9.39 & 6.52 & 6.50 & 7.70 & 6.34 & 6.39 & 37.03 & 28.54 & 27.14 \\
      \ding{55} & DAIR-V2X-I  & ImVoxelNet & 11.75 & 9.09 & 9.09 & 18.01 & 13.54 &  13.32 & 25.53 & 15.55 & 15.49 & 30.49 & 22.99 & 22.53 \\

        \rowcolor{gray!40}  \ding{55} & DAIR-V2X-I  & Only-Road & 24.31 & 13.54 & 13.17 & 31.89 & 18.42 & 16.69 & 44.24 & 27.46 & 25.32 & 47.60 & 28.42 & 28.22 \\ \hline 
        DAIR-V2X-V & DAIR-V2X-I & Addon & 26.74 & 15.36 & 13.84 & 33.97 & 18.83 & 18.51  & 46.07 & 27.36 & 27.01 & 49.81 & 30.51 & 28.25 \\ 
         DAIR-V2X-V & DAIR-V2X-I  & IROAM & \textbf{28.89} & \textbf{16.11} & \textbf{15.85} & \textbf{35.10} & \textbf{21.20} & \textbf{19.32} & \textbf{47.35} & \textbf{28.26} & \textbf{28.08} & \textbf{50.74} & \textbf{31.26} & \textbf{31.05} \\ \hline
         DAIR-Seq-V & DAIR-V2X-I  & Addon & 26.39 & 15.18 & 13.83 & 32.11 & 18.67 & 18.23 & 43.81 & 27.16 & 24.99 & 47.23 & 30.13 & 27.95 \\ 
         DAIR-Seq-V & DAIR-V2X-I  & IROAM & \textbf{27.48} & \textbf{16.05} & \textbf{15.70} & \textbf{35.08} & \textbf{21.09} & \textbf{19.26} & \textbf{46.96} & \textbf{28.02} & \textbf{25.76} & \textbf{50.43} & \textbf{30.98} & \textbf{28.68} \\ \hline \hline

        \rowcolor{gray!40}  \ding{55} & DAIR-Seq-I & Only-Road & 21.07 & 11.12 & 10.91 & 26.56 & 15.56 & 15.19 & 44.24 & 25.60 & 25.43 & 47.89 & 28.60 & 28.42 \\ 
         DAIR-V2X-V & DAIR-Seq-I   & Addon & 22.34 & 12.53 & 12.16 & 28.76 & 17.64 & 15.95 & 44.32 & 25.61 & 25.43 & 47.86 & 28.61 & 28.45 \\ 
         DAIR-V2X-V & DAIR-Seq-I   & IROAM & \textbf{24.51} & \textbf{13.66} & \textbf{13.49} & \textbf{31.82} & \textbf{18.50} & \textbf{16.73} & \textbf{44.56} & \textbf{27.79} & \textbf{25.64} & \textbf{48.03} & \textbf{28.74} & \textbf{28.62} \\ \hline
    \end{tabular}}
    \caption{Experimental results of roadside detection performance on DAIR-V2X \& DAIR-Seq.}
    \label{TAB:roadside_results}
\end{table*}

\textbf{Datasets and Metrics.} We conduct our experiments on DAIR-V2X~\cite{yu2022dairv2x}, which is a large-scale vehicle-infrastructure cooperative (VIC) dataset from real worl. The whole dataset contains 18330 roadside frames (called DAIR-V2X-I) and 20515 vehicle-side frames (called DAIR-V2X-V), both of which are divided into train/val/test sets in a 5:2:3 ratio. All experiments are evaluated on the val split. The resolution of images collected by RGB cameras is $1920\times1080$. Note that during training, we resize images to $1280\times720$ as input. Another VIC dataset, named V2X-Seq ~\cite{v2x-seq} is involved to verity its cross-domain learning ability. Following~\cite{v2x-seq}, vehicle-side part (DAIR-Seq-V) is split into train/val sets of 8504//3708 frames, and roadside part (DAIR-Seq-I) is divided into train/val sets of 7834/3441 frames.

The metrics for 3D object detection include three-level difficulties, i.e. easy, moderate, and hard (the same as KITTI~\cite{Geiger2013KITTI}). Each of them includes the average precision (AP) of
bounding boxes in 3D space and the BEV under a specific IoU threshold, denoted as $AP_{3D}$ and $AP_{BEV}$, which are all calculated under 40 recall positions.

\begin{table*}[ht]
  \footnotesize
  \centering
  \resizebox{\linewidth}{!}{
    \begin{tabular}{ccccccccccccccc}
    \hline
    \multirow{2}{*}{\underline{\textbf{Vehicle-side(Eval)}}} & \multirow{2}{*}{\textbf{Roadside}}& \multirow{2}{*}{\textbf{Method}} & \multicolumn{3}{c}{\textbf{$AP_{3D}(IoU=0.7)$}} & \multicolumn{3}{c}{\textbf{$AP_{BEV}(IoU=0.7)$}} & \multicolumn{3}{c}{\textbf{$AP_{3D}(IoU=0.5)$}} & \multicolumn{3}{c}{\textbf{$AP_{BEV}(IoU=0.5)$}}
     \\ \cline{4-15} 
     & & & Easy & Mod & Hard & Easy & Mod & Hard  & Easy & Mod & Hard & Easy & Mod & Hard \\ \hline
        \rowcolor{gray!40} DAIR-V2X-V & \ding{55}  & Only-Veh & 19.39 & 7.34 & 6.08 & 25.95 & 11.13 & 9.58 & 43.88 & 23.45 & 20.98 & 48.30 & 27.49 & 24.84 \\ 
         DAIR-V2X-V & DAIR-V2X-I & Addon & 26.48 & 15.28 & 13.71 & 32.07 & 18.57 & 18.16 & 43.38 & 26.95 & 24.78 & 48.91 & 30.03 & 27.82 \\ 
         DAIR-V2X-V & DAIR-V2X-I & IROAM & \textbf{27.29} & \textbf{15.94} & \textbf{15.49} & \textbf{35.17} & \textbf{21.10} & \textbf{19.29} & \textbf{47.05} & \textbf{28.10} & \textbf{27.89} & \textbf{50.48} & \textbf{31.10} & \textbf{30.91} \\ \hline

    \end{tabular}}
    \caption{Evaluation results of vehicle-side detection performance on DAIR-V2X.}
    \label{TAB:vehicle_results}
    \vspace{-0.5cm} 
\end{table*}

\begin{table*}[ht]
  \footnotesize
  \centering
    \begin{tabular}{ccccccccccccccc}
    \hline
    \multirow{2}{*}{\textbf{Veh Data}}& \multirow{2}{*}{\textbf{CL}}& \multirow{2}{*}{\textbf{DC}} & \multicolumn{3}{c}{\textbf{$AP_{3D}(IoU=0.7)$}} & \multicolumn{3}{c}{\textbf{$AP_{BEV}(IoU=0.7)$}} & \multicolumn{3}{c}{\textbf{$AP_{3D}(IoU=0.5)$}} & \multicolumn{3}{c}{\textbf{$AP_{BEV}(IoU=0.5)$}} \\ \cline{4-15} 
     & & & Easy & Mod & Hard & Easy & Mod & Hard & Easy & Mod & Hard & Easy & Mod & Hard \\ \hline
     & & & 24.31 & 13.54 & 13.17 & 31.89 & 18.42 & 16.69 & 44.24 & 27.46 & 25.32 & 47.60 & 28.42 & 28.22 \\ 
     \checkmark & & & 24.61 & 13.63 & 13.27 & 29.61 & 16.39 & 16.13 & 43.62 & 25.13 & 24.90 & 47.08 & 28.05 & 27.77 \\
       & \checkmark & & 26.72 & 15.53 & 14.00 & 34.99 & 21.10 & 19.21 & 44.64 & 27.75 & 25.55 & 47.83 & 28.58 & 28.36 \\
     \checkmark & \checkmark & & 26.90 & 15.73 & 15.25 & 34.60 & 19.27 & 18.94 & 46.40 & 27.64 & 25.44 & 49.94 & 30.69 & 28.39 \\
     \checkmark & \checkmark & \checkmark & \textbf{28.89} & \textbf{16.11} & \textbf{15.85} & \textbf{35.10} & \textbf{21.20} & \textbf{19.32} & \textbf{47.35} & \textbf{28.26} & \textbf{28.08} & \textbf{50.74} & \textbf{31.26} & \textbf{31.05} \\ \hline
    \end{tabular}
    \caption{Ablation study on each component of IROAM.  CL means contrastive learning and DC means semantic-geometry decouple.
    }
    \label{TAB:AB}
\end{table*}

\textbf{Training.} Before training, we use random crop for data augmentation following~\cite{zhang2022monodetr}.
During training, we ignore the samples with depth labels larger than 65 meters or smaller than 2 meters and adopt the weighted values $\lambda_{1 \sim 7}$ as 2, 10, 5, 2, 1, 1, 1 respectively.
During inference, we set the threshold of 0.2 on category confidence to filter out the object queries without NMS post-processing. We set the number of queries $N$ as 50 and the channel dimension of queries as 256. We train the module for 195 epochs on 2 NVIDIA A30 GPUs with batch size 6 and the learning rate $5\times10^{-5}$. We adopt AdamW optimizer with weight decay $10^{-4}$ and decrease the learning rate at 125 and 165 epochs by a factor of 0.1.

\subsection{Object Detection Performance}

We conduct three groups of experiments(Line 1 to 7 in Table~\ref{TAB:roadside_results}) based on DAIR-V2X-I datasets to evaluate what extent of detection improvement can be brought by IROAM. The Only-Road model removes Cross-Domain Query Enhancement from IROAM, which will be degraded as a monocular detector. Only-Road outperforms BEVHeight and ImVoxelNet and all of them use only roadside data for training. We conduct the other two groups of experiments, one utilizes DAIR-V2X-V as the source of vehicle-side data and the other uses DAIR-Seq-V. Only-Road only uses roadside data for training and Addon blends vehicle-side and roadside data.  The results show that Addon has a better performance than Only-Road, which demonstrates that vehicle-side data can help learn a better representation of roadside data for 3D object detection. IROAM is only trained with pairs of vehicle-roadside images and increases the improvement to 18.84\% /18.98\% /20.34\% for $AP_{3D}(IoU =0.7)$ metrics of Easy/Moderate/Hard parts compared with Only-Road method.

We select DAIR-Seq-V as vehicle-side data and DAIR-V2X-I as roadside data to validate the model's generalization. The results show that IROAM can also bring great performance improvement on all $AP_{3D}$ and $AP_{BEV}$. We conduct another group of experiments, which select DAIR-V2X-V as vehicle-side data and DAIR-Seq-I as roadside data. IROAM also has better performance than Only-Road and Addon. Experiments indicate that IROAM can learn information across vehicle-side and roadside data domains.

Experiments in Table~\ref{TAB:vehicle_results} also show that IROAM can also be beneficial to the detection of vehicle-side domain. Addon can bring the detection gain on vehicle-side data by introducing roadside data compared with Only-Veh model (only uses vehicle-side data for training). IROAM can achieve 0.81 /0.66 /1.78 performance gain for $AP_{3D}(IoU =0.7)$ of Easy/Moderate/Hard parts compared to Addon method.

\begin{figure}[htbp]
	\centering  
	\includegraphics[width=0.8\linewidth]{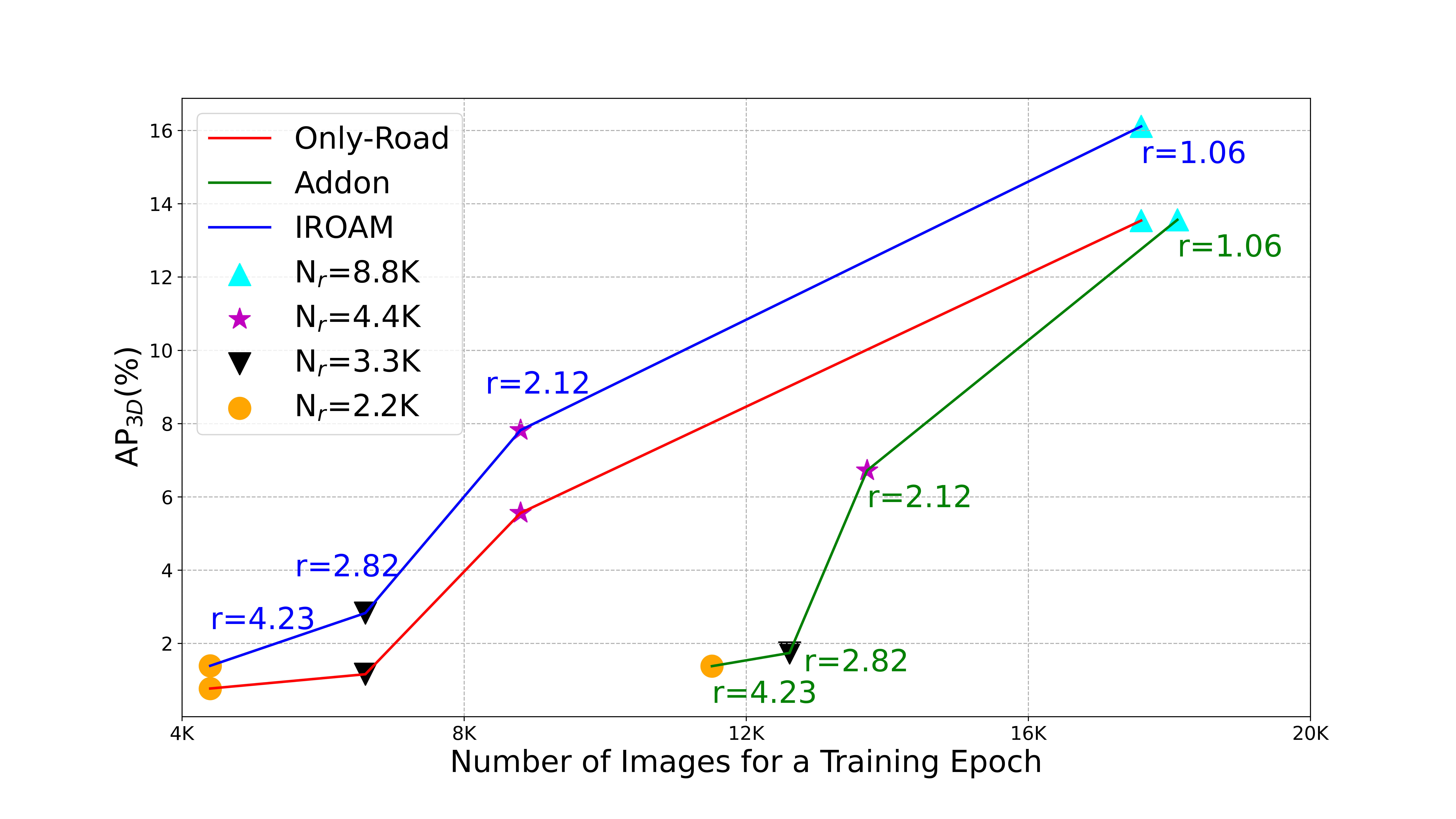} 
	\caption{Analysis on different proportion ratios of vehicle-side to roadside data. Four groups of experiments (Using different icons to distinguish them) simulate a significant imbalance between roadside and vehicle-side data, which select $N_{r} = 2.2/3.3/4.4/8.8\text{K}$ roadside images respectively. Since Addon uses all vehicle-side data in one epoch, the ratio of vehicle-side data to roadside data $r$ is 4.23/2.82/2.12/1.06. IROAM samples partial images from all vehicle-side images to form pairs with roadside images for training. Only-Road repeats roadside data twice within one epoch for a fair comparison so that $r$ is always 1.0. The y-axis means the $AP_{3D}(IoU =0.7)$ for Moderate parts and the x-axis means the number of images used in one training epoch.}  
	\label{fig:ratio}   
\end{figure}

\subsection{Ablation Studies}

We regard the Only-Road model (first line in Table~\ref{TAB:AB}) as the baseline in the ablation study, where all experiments are evaluated on DAIR-V2X-I dataset. Veh Data means to train the model with both roadside and vehicle data. CL means adding contrastive learning loss to the final loss. DC means semantic-geometry decouple operation as depicted in Section~\ref{Section:CDQE} . Results show that learning from vehicle dataset and conducting CL on two sides of data can help improve detection performance on roadside. DC can increase {$AP_{3D}(IoU=0.7)$ 1.99 compared with only using CL (Compare the 4th and 5th row). This phenomenon can validate that our proposed decouple operation can help IROAM focus on the similarity between two data domains and learn a better representation for detection through cross-domain interaction.

\subsection{Analysis on Imbalance between Two-side Data}

In the real world, there is a vast amount of vehicle-side data, which is several orders of magnitude larger than roadside data. DAIR-V2X-C dataset has 8800 frames of roadside images and 9314 frames of vehicle-side images for training, which is roughly equal for two sides. To simulate the scenario that vehicle-side data is greatly larger than roadside data, we combine partial roadside data with whole vehicle-side data for training to simulate such an imbalance between two-side data. We conduct four groups of experiments on Only-Road, Addon and IROAM model, which select $N_{r} = 2.2/3.3/4.4/8.8\text{K}$ roadside images for training respectively. Three models learn from all vehicle-side images but with different training pipelines so that different models sample different numbers of images in one training epoch.


The results are depicted in Figure~\ref{fig:ratio}. Although Addon can increase $AP_{3D}$ compared with Only-Road in every group, it needs to traverse all vehicle-side and roadside images in a training epoch, which brings a large overhead. IROAM can improve detection performance over Only-Road in all groups without bringing any extra calculation operation and can work well when there is a significant imbalance between roadside and vehicle-side data.

\section{CONCLUSIONS}
The roadside model can bring a holistic view of the surrounding environment for autonomous vehicles, but it needs some extra roadside images for training a new detector, which is time-consuming to capture and annotate. To tackle this challenge, we proposed IROAM, a semantic-geometry decoupled contrastive learning framework, which leverages tremendous vehicle-side data to enhance roadside object detection. We first extract object queries from two data domains in In-Domain Query Interaction. We decouple object queries with a semantic part and a geometry part and only the former is utilized to calculate the contrastive loss between two data domains in Cross-Domain Query Enhancement. Experiments conducted on various vehicle-side and roadside datasets demonstrate the effectiveness of IROAM in improving roadside object detection performance.

\section*{ACKNOWLEDGMENT}
This work is funded by the National Science and Technology Major Project (2022ZD0115502) and Lenovo Research.









\bibliographystyle{IEEEtran}
\balance
\bibliography{IEEEexample}

\end{document}